# Qualitative Differences Between Evolutionary Strategies and Reinforcement Learning Methods for Control of Autonomous Agents

**Nicola Milano & Stefano Nolfi**

Laboratory of Autonomous Robots and Artificial Life,
Institute of Cognitive Science and Technologies, National Research Council, Roma, Italy
nicola.milano@istc.cnr.it, stefano.nolfi@istc.cnr.it

**Abstract.** In this paper we analyze the qualitative differences between evolutionary strategies and reinforcement learning algorithms by focusing on two popular state-of-the-art algorithms: the OpenAI-ES evolutionary strategy and the Proximal Policy Optimization (PPO) reinforcement learning algorithm --- the most similar methods of the two families. We analyze how the methods differ with respect to: (i) general efficacy, (ii) ability to cope with sparse rewards, (iii) propensity/capacity to discover minimal solutions, (iv) dependency on reward shaping, and (v) ability to cope with variations of the environmental conditions. The analysis of the performance and of the behavioral strategies displayed by the agents trained with the two methods on benchmark problems enable us to demonstrate qualitative differences which were not identified in previous studies, to identify the relative weakness of the two methods, and to propose ways to ameliorate some of those weakness. We show that the characteristics of the reward function has a strong impact which vary qualitatively not only for the OpenAI-ES and the PPO but also for alternative reinforcement learning algorithms, thus demonstrating the importance of optimizing the characteristic of the reward function to the algorithm used.

## 1. Introduction

Evolutionary strategies (ES) and reinforcement learning (RL) algorithms represent two well-established techniques for training embodied and situated agents. Both methods permit to train agents from scratch on the basis of a fitness or reward function which rates how well the agent is behaving.

In this article we analyze the qualitative difference between the two methods. We discuss in particular the following aspects: (i) the general efficacy, (ii), the ability to cope with sparse rewards, (iii) the propensity/capacity to discover minimal solutions, (iv) the dependency on reward shaping, and (v) the ability to cope with variations of the environmental conditions. For the first two aspects we will briefly discuss the evidences available in the literature. For the remaining three aspects, we provide original experimental evidence and we discuss the causing factors.

Clarifying the qualitative differences permits to choose the most suitable algorithm and hyperparameters in a more informed way. Moreover, it can permit to identify how to mitigate the weakness of the algorithm chosen.

Clearly, carrying out a systematic comparison of all existing algorithms on a large set of benchmark problems is out of the scope of a single paper. This also in consideration of the fact that identifying qualitative differences requires in depth studies which go behind the mere comparison of performance measures. Consequently, our analysis does not aim to be exhaustive. The objective of this paper is that to illustrate qualitative differences which have not been analyzed in previous studies and to start collecting experimental data which can be used to characterize some of the existing algorithms with respect to those differences.

For our analysis we focus primarily on two popular state-of-the-art algorithms: the OpenAI-ES evolutionary strategy (Salimans et al., 2017) and the Proximal policy optimization reinforcement learning algorithm (PPO, Schulman et al., 2017). We start from these algorithms also in consideration of the fact that they are the most similar among the evolutionary and reinforcement learning families. Indeed, they both operate on-policy and they both optimize a gradient on the basis of the Adam stochastic optimizer (other evolutionary algorithms operate without estimating a gradient and without using a stochastic optimizer). However, for some of the analysis we also consider two off-policy reinforcement learning algorithms: the T3D (Fujimoto et al., 2018) and the Soft Actor-Critic (SAC) (Haarnoja et al. 2018). As we will see, the analysis of those algorithms also reveals interesting qualitative differences among reinforcement learning algorithms.

The OpenAI-ES algorithm belongs to the class of natural evolution strategies (NES) (Wierstra et al., 2008, 2014 , Sehnke et al., 2011, Glasmachers et al. 2010, Schaul et al. 2011). Let F denote the objective function acting on parameters θ, the algorithm generates a population with a Gaussian distribution over θ parametrized by ψ and tries to maximize the mean objective value $E_{\theta \sim p_\psi} F(\theta)$ over the population by using the Adam stochastic optimizer. During each iteration it takes a gradient step on ψ by using the following estimator:

$$\nabla_\psi E_{\theta \sim p_\psi} F(\theta) = E_{\theta \sim p_\psi}[F(\theta) \nabla_\psi \log p_\psi(\theta)] \quad (1)$$

The Proximal Policy Optimization Algorithm (PPO) is an actor-critic on-policy reinforcement learning method (Mnih et al 2016, Zhang et al. 2018, Konda et al. 1999, Haarnoja et al. 2018) which uses a parametrized stochastic policy. It operates on the basis of a surrogate gradient that penalizes excessive divergence from the previous policy by clipping the gradient in a proximal trusted zone of the search space.

It updates the policy by using the Adam stochastic optimizer via:

$$\theta_{k+1} = \arg\max_\theta \mathop{E}_{s,a \sim \pi_{\theta_k}} [L(s, a, \theta_k, \theta)] \quad (2)$$

where *s* are the observation states, *a* the actions taken, $\theta$ are the parameters of the network policy and L is the loss defined as:

$$L(s, a, \theta_k, \theta) = \min\left(\frac{\pi_\theta(a \mid s)}{\pi_{\theta_k}(a \mid s)} A^{\pi_{\theta_k}}(s, a), \text{clip}\left(\frac{\pi_\theta(a \mid s)}{\pi_{\theta_k}(a \mid s)}, 1-\epsilon, +\epsilon\right) A^{\pi_{\theta_k}}(s, a)\right) \quad (3)$$

where $\frac{\pi_\theta(a|s)}{\pi_{\theta_k}(a|s)}$ is the probability ratio of performing those actions with the current and the previous version of the policy, $A^{\pi_{\theta_k}}(s, a)$ is the advantage, and $\epsilon$ is the threshold used to clip the gradient.

The Twin Delayed DDPG (TD3) is an off-policy reinforcement learning method which operates with a deterministic policy combined with two Q-functions. The usage of the smaller Q-value computed by the two Q-functions permit to reduce the problems caused by the overestimation of the Q-values. The algorithm learns the control policy $\pi_\theta$ and two Q-functions $Q_{\phi_1}$, $Q_{\phi_2}$ in an alternate way by mean square error Bellman minimization:

.

$$y(r, s', d) = r + \gamma(1-d)\min_{i=1,2} Q_{\phi_{i,\text{targ}}}(s', a'(s')) \quad (4)$$

and then both are learned by regressing to this target:

$$L(\phi_i, \mathcal{D}) = \mathop{E}_{(s,a,r,s',d) \sim \mathcal{D}}\left[\left(Q_{\phi_i}(s, a) - y(r, s', d)\right)^2\right], \ i = 1,2 \quad (5)$$

Moreover, it uses a second policy network which contains a time-delayed version of the parameters of the first network to eliminate the instabilities caused by the need to optimize a target that depends on the same parameters to be optimized.

Like other algorithms which approximate the Q function, it uses an experience reply buffer which contains previous experiences. The policy is learned by maximizing $Q_{\phi_1}$.

The Soft Actor-Critic (SAC) algorithm is similar to the TD3 algorithm. However, it uses a stochastic policy instead than a deterministic policy. Moreover, it determines the next state on the basis of the current policy instead than on the basis of a previous version of the policy.

The policy is trained to maximize a trade-off between the expected return and entropy, a measure of randomness of the policy.

$$\pi^* = \arg\max_\pi \mathop{E}_{\tau \sim \pi}\left[\sum_{t=0}^\infty \gamma^t \left(R(s_t, a_t, s_{t+1}) + \alpha H(\pi(\cdot \mid s_t))\right)\right]$$

Where H is the entropy of the policy computed from its distribution and $\alpha$ the trade-off coefficient.

The experiments reported in this article can be replicated by using evorobotpy2, available from https://github.com/snolfi/evorobotpy2, and stable baseline available from https://github.com/hill-a/stable-baselines. The modified reward functions described in Section 5 are implemented with evorobotpy2 and can be used by specifying the version 5 of the problems (e.g. HopperBulletEnv-v5, in the case of the Pybullet Hopper problem).

## 2. General efficacy

Comparing the general efficacy of EA and RL is far from trivial since the performance depends on the problems considered, on the setting of the hyperparameters, and on the fitness or reward functions used, as we will also illustrate in the following sections.

In any case, overall, the existing comparative studies do not provide evidence indicating a general superiority of one class of methods over the other. Given the limited number of studies which compared the OpenAI-ES and the PPO we discuss here also the studies carried out with the CMA-ES (Hansen & Ostermeier, 2001) another state-of-the-art evolutionary algorithm similar to the OpenAI-ES and the TRPO (Schulman et al., 2015), an algorithm similar to the PPO.

The results reported in Duan et al. (2016) obtained with the CMA-ES and TRPO indicate a general superiority of the latter over the former. Indeed, the TRPO outperforms the CMA-ES on 11 out of 12 classic control and MuJoCo locomotors problems. On the other hand, this result is biased from the fact that the duration of the training is determined on the basis of the total number of evaluation steps without considering that the computation cost of the TRPO algorithm is higher than that of the CMA-ES algorithm. Evolutionary algorithms tend to be less sample efficient than reinforcement learning methods in general but are less computationally expensive and can benefit more from parallelization. Indeed, in a subsequent work Salimans et al. (2017) demonstrated that the OpenAI-ES evolutionary algorithm and the TRPO reinforcement learning algorithm achieve similar performance on the MuJoCo locomotors problems and on the Atari problems at equal computational cost. Moreover, they demonstrated that highly parallel implementations of evolutionary algorithms permit to generate solutions in a remarkably short period of time.

The dependency of the relative performance on the problem considered can be appreciated in the study of Zhang and Zaiane (2017) who observed the superiority of one type of algorithm in certain problems and of the other type in the remaining problems.

The impact of the fitness or reward function is discussed in Section 5.

## 3. Ability to cope with sparse reward

Another property which differentiates evolutionary and reinforcement learning is the ability to cope with sparse rewards, i.e. rewards obtained after a sequence of actions.

EAs do not suffer from the sparsity of the reward since they operate on the basis of a fitness measure that encodes the sum of the rewards collected during evaluation episodes. RLs instead, which operate by associating rewards to specific actions, struggle with temporal credit assignment when rewards are sparse. Temporal difference in RL use bootstrapping to better handle this aspect but still struggles with sparse rewards when the time horizon is long.

We do not analyze this qualitative difference in more detail in this paper since it is well documented in the literature (see for example Khadka & Tumer (2018), Zhu, Belardinelli & León B. G. (2021)).

## 4. Propensity/capacity to discover minimal solutions

Both evolutionary and reinforcement learning algorithms introduce random variations to discover better policies. However, they differ in the way in which variations are introduced. Indeed, the OpenAI-ES algorithm, and more generally evolutionary algorithms, introduce variations in the parameters of the policy during the generation of a new population of policies while the PPO algorithm, and most reinforcement learning algorithms, introduce variations in the actions performed by the policy in each step (for alternative reinforcement learning methods which introduce variations in the parameters of the policy see [Plappert et al. 2017 and Raffin and Stulp 2020]). Moreover, in the PPO algorithm the distribution of variations is encoded in adapting parameters, is initially high and usually decreases during the course of the training. In the OpenAI-ES algorithm, instead, the distribution of variation is constant. These differences have important consequences.

A first consequence is that the set of solutions which are available to the two algorithms differ. By using deterministic or almost-deterministic policies, the OpenAI-ES algorithm can adopt simple solutions, i.e. solutions which permit to solve the problem on the basis of few simple control rules. These simple solutions cannot be realized through the usage of stochastic policies and consequently cannot be selected by the PPO algorithm.

This difference can represent an advantage or a disadvantage depending on the circumstances. In the case of problems which admit simple solutions which are optimal, it represents an advantage for the OpenAI-ES algorithm and a disadvantage for the PPO algorithm, which is forced to find more complex solutions. In the case of problems which admit only suboptimal simple solutions, it represents a disadvantage for the OpenAI-ES algorithm, which risk to remain trapped into a suboptimal solution, and an advantage for the PPO algorithm, which does not have access to those simple solutions.

This difference can be illustrated by analyzing the solutions ons discovered by the OpenAI-ES and PPO algorithm for the Centipede and Breakout Atari problems. The OpenAI-ES algorithm solves these two problems by moving the player in a specific location and by later keeping the player in that precise position (see Figure 1 center and Videos 1-3. The links to the videos are included in the Appendix). This minimal strategy enables the agent to collect a very high reward in some problems, like the Centipede game, but is suboptimal in other problems like the Breakout game. This strategy is clearly inaccessible for an agent trained with the PPO which is exposed to strong action perturbations. Consequently, the PPO is forced to develop more complex strategies which rely on the ability to intercept the ball from different positions (Figure 1, right and Video 2-4). In the case of the Breakout problem, the possibility of the OpenAI-ES algorithm to select a minimal strategy is counter-productive since the strategy is suboptimal and constitutes a local minimum, i.e. prevent the possibilities to later select better strategies. Consequently, the PPO outperforms the OpenAI-ES algorithm on this problem. In the case of the Centipede problem, instead, the possibility to select this minimal strategy is advantageous since the strategy is more effective than alternative strategies. Consequently, the OpenAI-ES algorithm outperforms the PPO in this problem.

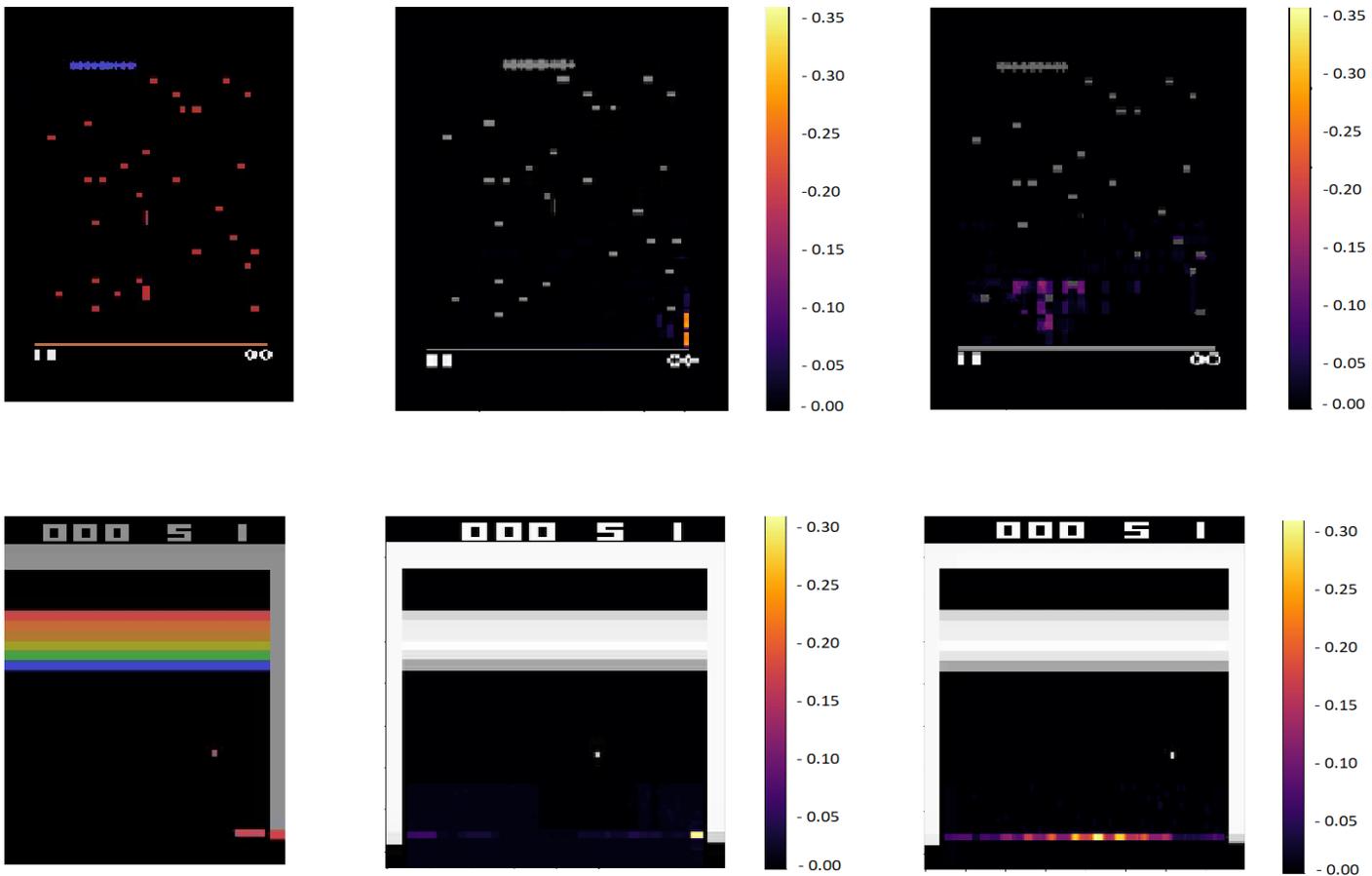

Figure 1. Positions assumed by a typical player trained on the Centipede and Breakout Atari problems (top and bottom, respectively). The first column shows a screenshot of the game in color. The other columns show in color at the bottom the frequency with which the agent is situated in different locations at the bottom during the game. The second and third columns display the data of the agents trained with OpenAI-ES and PPO algorithms, respectively, Data obtained by running 5 replications of each experiment. As can be seen, the agents trained with the PPO algorithm move in different locations during the game while the agents trained with the OpenAI-ES algorithm remain in the same location for most of the game. Hyperparameters are reported in the appendix.

The ability of evolutionary algorithms to discover simple and clever strategies is well-documented in the literature (Lehman et al., 2020; Wilson et al. 2018). As far as we know, however, the fact that some of the strategies available to evolutionary algorithms are not available to reinforcement learning algorithms was not discussed before.

The qualitative difference between the OpenAI-ES and the PPO can be reduced by perturbing the actions also in the case of the former algorithm. This technique is usually used to evolve solutions which are robust to environmental variations (Pagliuca and Nolfi, 2019) and/or to evolve solutions which can cross the reality gap, i.e. which keep operating effectively once they are moved from the simulated to the real environment (Jakobi, Husbands, & Harvey, 1995).

Figure 2 displays the performance achieved by training deterministic policies, stochastic policies subjected to minor perturbations and stochastic policies subject to much larger perturbations with the OpenAI-ES algorithm. In the former cases the perturbations are generated with the addition of Gaussian random values with an average of 0.0 and standard deviation of 0.01. In the latter case, the distribution of perturbation for each action value is parametric and is initially set to values close to 1.0 (as in the case of the policies used by the PPO). As can be seen, the addition of small perturbations to the action vector permits to generate solutions which are robust with respect to this form of perturbation. Moreover, it leads to better performance in part of the problems considered. On the other hand, the utilization of the parametric gaussian policies used by the PPO leads to significantly worse performance in all cases.

Overall, this implies that the OpenAI-ES algorithm tolerates and can even benefit from the addition of small action perturbations. However, unlike the PPO, it is unable to tolerate large perturbations.

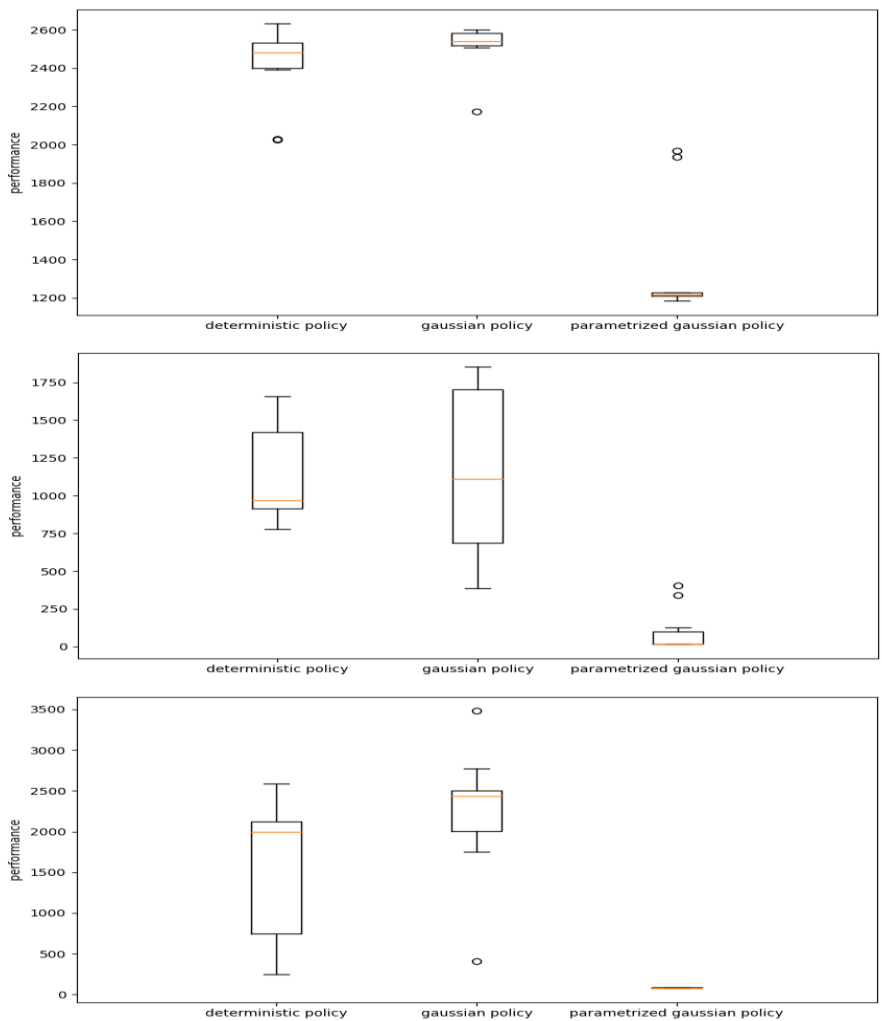

**Figure 2. Maximum performance obtained with the OpenAI-ES algorithm on the Pybullet (Coumans & Bai, 2016) Hopper (up), Walker2d (center) and Humanoid (bottom) environments. The boxplots display the performance obtained by using a deterministic policy, a gaussian policy in which actions are perturbed with tiny random values (i.e. random values generated with an average of 0.0 and a distribution of 0.01), and a parametrized diagonal gaussian policy. The second experimental condition produces significantly better performance than the first experimental condition in the case of the Hopper and Humanoid problems (Wilcoxon non parametric test p-value < 0.001) and equally good performance in the case of the Walker2D problem (Wilcoxon non parametric test p-value > 0.05). The third experimental conditions produce worse performance than the**

first experimental condition in all cases (Wilcoxon non parametric test p-value < 0.001). Hyperparameters are reported in the appendix.

## 5. Dependency on reward shaping

Evolutionary and reinforcement learning algorithms also differ with respect to their dependency on reward shaping. Reward shaping refers to the attempt to facilitate the development of effective solutions by designing reward functions which rate the learning agents not only for their ability to solve the task but also for abilities which are instrumental to the solution of the problem. For example, the development of a walking ability can be obtained by rewarding the agent on the basis of its speed toward the destination only, i.e. on the basis of its ability to walk fast only. Alternatively, it can be obtained by also rewarding the learning agent for the ability to remain upright. This additional reward component (incentive) is intended to favor the development of an ability to stay upright which does not constitute a solution of the problem by itself but which can facilitate the development of the ability to walk. Unfortunately, however, identifying useful incentives can be challenging since the effect of the incentives is hard to predict and since the introduction of incentives can promote the development of non-effective solutions which maximize the incentives without maximizing the primary component of the reward function (Andrychowicz at al., 1999; Nolfi, 2021).

Many benchmark problems include incentives in their reward functions. For example, the Pybullet locomotors problems and the Bipedal hardcore problem include a bonus for staying upright. In this section we analyze the effect that the incentive has on agents trained on the Hopper and Walker2d Pybullet problems and on the Bipedal Hardcore problem with the OpenAI-ES algorithm and with the PPO, T3D and SAC reinforcement learning algorithms.

The comparison of the performance obtained with and without the incentive (Figure 3 and 4, respectively) and the visual inspection of the behavior of the trained agents indicates that the incentive has a strong impact on the results. The impact is positive or negative depending on the problem and on the algorithm used. Overall, the addition of the incentive has a positive impact on the cases in which the agents do not manage to develop a walking ability without the incentive and a negative impact on the cases in which the agents are unable to develop a walking ability without the incentive.

In the case of the Bipedal Hardcore problem, the agents manage to develop an ability to walk without the incentive with all algorithms. Consequently, the distance travelled by the agents trained with the incentive is lower in all cases.

For the other two problems, the OpenAI-ES algorithm manage to develop a walking behavior without the incentive and fail to walk with the incentive. This confirms that the addition of the incentive is counterproductive in settings in which the agents manage to solve the task also without it (see also Pagliuca, Milano & Nolfi, 2021). The null performance obtained with the incentive is due to the fact that in this experimental condition the agents develop an ability to avoid falling by staying still which maximize the reward obtained through the incentive only. The PPO algorithm is unable to develop a walking ability without the incentive (i.e. the trained agents just fall down after few steps) and benefit from the incentive. The TD3 also benefit from the incentive since it fails to develop an ability to walk in the case of the Walker2D problem and in some of the replications of the Hopper problem. The SAC algorithm manages to produce a walking behavior without the incentive in most of the replications and produce similar performance on the average with the incentive. The fact that the distribution of performance among replications is wider without the incentive and smaller with the incentive, suggests that in the case of SAC the incentive has a positive impact on the replications which fail producing a walking behavior and a negative impact on the other replications.

The differences among the algorithms are probably due to usage of deterministic versus stochastic policies and to differences in the exploration abilities. The usage of policy which are highly stochastic, especially during the initial phase of the training process, increases the complexity of the problem to be solved since it requires to discover a set of parameters which enable the agent to walk in the presence of strong action perturbations. Consequently, the usage of stochastic policy increases the potential utility of the incentive. This consideration can be used to explain why the PPO algorithm, which use a stochastic policy, is completely unable to discover a walking ability in the case of the Hopper and Walker2D problems without the incentive and benefit from the incentive. The results obtained with the TD3 and SAC algorithm, on the other hand, seem to imply that the algorithms differ also with respect to the ability to explore the search space independently from whether they use a stochastic or deterministic policy. This is not surprising since evolutionary and reinforcement learning method differ with respect to the way in which they introduce variations and since the SAC algorithm, which is less dependent on the incentive, improves its exploration ability through the maximization of entropy.

More generally, the results reported in this section demonstrates that the performance of all algorithms are dramatically affected from the characteristics of the reward functions and that reward functions optimized for an algorithm can be very sub-optimal for other algorithms. This implies that benchmarking alternative methods without optimizing the reward function can be useless, an issue which has been neglected to date.

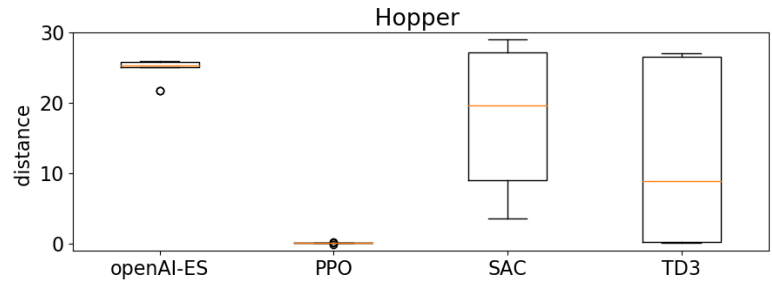

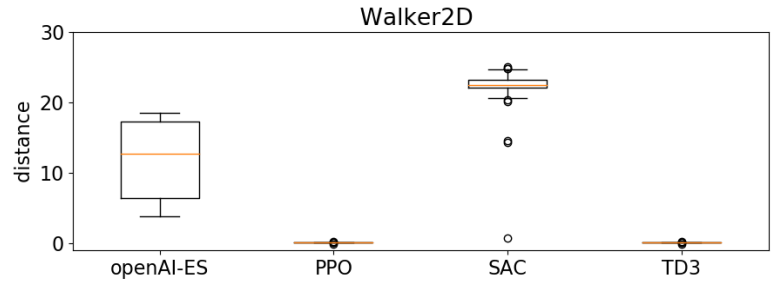

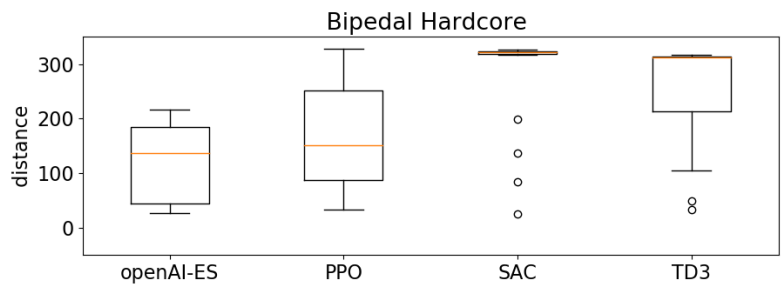

**Figure 3.** Average distance traveled by Hopper, Walker2D, and Bipedal Hardcore agents trained with the OpenAI-ES, PPO, TD3 and SAC algorithm. Results obtained with the reward function which does not include the incentive for remaining upright. Each boxplot shows the results obtained post-evaluating for 5 episodes the best 10 trained agents of the 10 corresponding replications. Boxes represent the inter-quartile range of the data and horizontal lines inside the boxes mark the median values. The whiskers extend to the most extreme data points within 1.5 times the inter-quartile range from the box. The hyperparameters used are described in the Appendix.

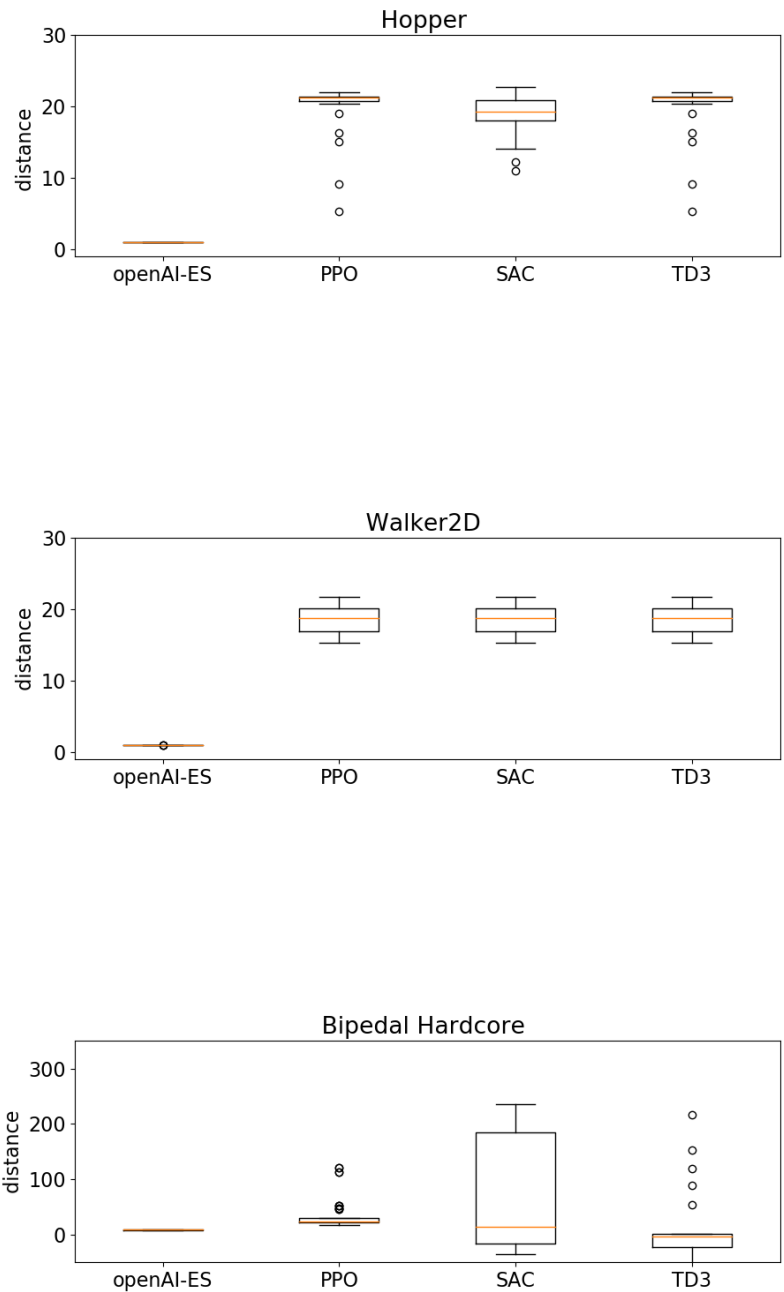

**Figure 4.** Average distance traveled by Hopper, Walker2D, and Bipedal Hardcore agents trained with the OpenAI-ES, PPO, TD3 and SAC algorithm. Results obtained with the standard reward function which includes an incentive for remaining upright. Each boxplot shows the results obtained post-evaluating for 5 episodes the best 10 trained agents of the 10 corresponding replications. Boxes represent the inter-quartile range of the data and horizontal lines inside the boxes mark the median values. The whiskers extend to the most extreme data points within 1.5 times the inter-quartile range from the box. The hyperparameters used are described in the Appendix.

# 6. Ability to cope with environmental variations

Still another property which differentiates EAs and RLs is the ability to cope with environmental variations. Exposing the agents to variable environmental conditions is necessary to promote the development of solutions which are robust with respect to variations, i.e. to avoid the selection of solutions overfitted to the specific environmental conditions experienced during the evolutionary or learning process (Pagliuca & Nolfi, 2019). The environmental conditions can be varied by changing the initial position/orientation of the robot and of the objects present in the environment at the beginning of evaluation episodes and by perturbing the state of the robot and/or the environment during the episode. The introduction of environmental variations, however, makes the reward measure noisy. Indeed, in the presence of environmental variations, the fitness or the reward collected by an agent does not depend on the skill of the agent only but also on the environmental conditions encountered during the agent's evaluation. The fitness or reward will thus be overestimated or underestimated for agents which encountered favorable or unfavorable environmental conditions, respectively. Environmental variations thus reduce the information contained in the reward measure. The noise affecting the reward measure can be reduced by evaluating the agents for multiple episodes. However, the number of episodes should be kept small to avoid an excessive increase of the computation cost.

Reinforcement learning methods, like the PPO, can tolerate greater environmental variations than evolutionary methods such as the OpenAI-ES. This can be explained by considering that the latter algorithm, and more generally ESs, operate by introducing variations producing high fitness values while the PPO, and many other RLs, operate by introducing variations producing positive advantages, i.e. to reward values which are higher than those expected in the given agent/environmental state. The expected reward is higher and smaller in easy and difficult environmental conditions, respectively. Consequently, the utilization of the advantage permits to filter out from the reward measure the effects of the variations of the environmental conditions, providing that the prediction of the expected reward is accurate.

The fact that the PPO algorithm can tolerate higher environmental variations than the OpenAI-ES algorithm can be illustrated with the Slime Volley environment (Ha, 2020). This is a volley game in which two agents are situated in the two subparts of a field divided by a net. The goal of each agent is to send the ball on the ground of the opponent and avoid the ball touching its own ground. In the version of the problem considered, the left agent is trained while the right agent is provided with a pre-trained policy which remains constant. In a problem of this kind the orientation and velocity with which the ball is launched at the beginning of the episode has a strong impact on the fitness/reward collected by the agent, especially at the beginning of the evolutionary or learning process in which the ability of the agent to intercept the ball is poor. In particular, at the beginning of the process the reward obtained by an agent correlates primarily with the fraction of episodes in which the ball is launched toward the agent's own field and with the initial orientation and velocity with which the ball is launched --- two factors which are independent from the agent's skill.

The fact that the PPO tolerates the effect of those environmental variations better than the OpenAI-ES algorithm is demonstrated by the fact the PPO manages to quickly develop effective agents while the OpenAI-ES fails (Figure 5, top). Moreover, it is demonstrated by the fact that the OpenAI-ES algorithm manages to solve the symmetrical version of the Slime Volley problem in which the noise caused by the environmental variation is substantially reduced without reducing the overall range of variation of the environmental conditions (Figure 5, bottom). In the symmetrical version of the problem, the orientation and velocity of the ball is generated randomly during even episodes while is generated by inverting the angle of 180 degrees and by maintaining the same velocity of the previous episode, during odd episodes. This ensures that the number of times in which the ball is launched in the field of the two players and the relative angle with which it is launched are the same. The modification of the problem introduced thus reduces the impact of the environmental variations without altering the complexity of the problem to be faced and the distribution of conditions to be faced.

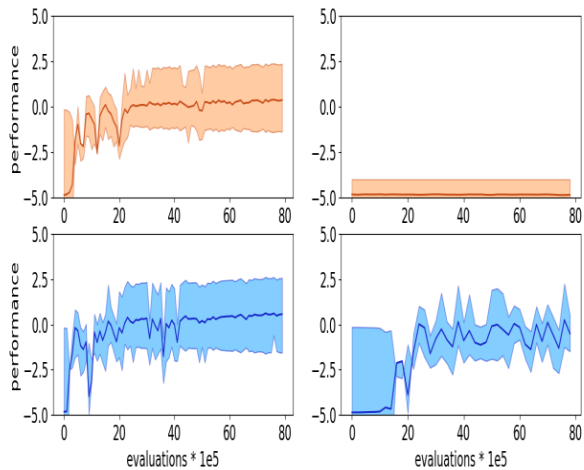

**Figure 5.** Performance obtained by agents trained with the PPO and OpenAI-ES algorithm, left and right respectively, during the training process. Data obtained on the standard and symmetrical versions of the Slime Volley problem, top and bottom respectively. Mean and 90% bootstrapped confidence intervals of the mean (shadow area) across 10 replications per experiment.

The weakness of the OpenAI-ES algorithm in that respect can be reduced by using the super-symmetrical variation of the original algorithm introduced here in which symmetrical individuals are exposed to the same environmental conditions (Figure 6). The algorithm uses symmetrical sampling to improve the accuracy of the estimated fitness gradient. This means that the population is formed by couples of offspring which are generated by perturbing the parameters of the parent through the addition and subtraction of the same vector of random values. In the standard version of the algorithm, each individual is exposed to randomly different environmental conditions. In the super-symmetrical version of the algorithm introduced here, inste
ad, each couple of symmetrical offspring is exposed to randomly different environmental conditions but the two symmetrical offspring are exposed to the same environmental conditions. This permits to estimate the relation between the perturbation received by each couple of symmetrical offspring and the fitness independently from the effect of environmental variations. As shown by Figure 6, this method permits to achieve significantly better performance than the standard OpenAI-ES algorithm in the modified version of the task (Wilcoxon non parametric test p-value < 0.01)

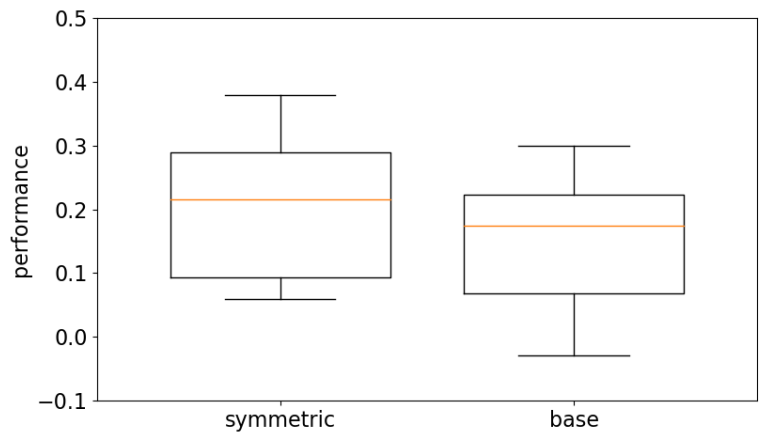

**Figure 6.** Performance obtained by agents trained with the super-symmetrical version of the OpenAI-ES algorithm (super-symmetrical) and with the standard version of OpenAI-ES algorithm (standard) on the symmetrical version of the Slime Volley problem, average results of best individuals over 10 replications of the experiment. Each boxplot shows the results obtained in 10 replications. Boxes represent the inter-quartile range of the data and horizontal lines inside the boxes mark the median values. The whiskers extend to the most extreme data points within 1.5 times the inter-quartile range from the box.

Another method which can be used to reduce the noise of the fitness measure in evolutionary algorithms consists in estimating the relative difficulty of environmental conditions and choosing conditions which have similar levels of difficulty on the average (see Milano and Nolfi, 2021).

## 7. Discussion

In this article we compared the EAs and RLs by focusing in particular on the OpenAI-ES and PPO algorithms which are the most similar methods belonging to the two classes and which represent the state-of-the-art in their respective class. The results of the comparisons reported in the literature and the original results reported here do not indicate a general superiority of one algorithm over the other but rather that the two methods differ qualitatively with respect to several factors. The appreciation of these factors can be crucial to identify the method which is most promising for a particular problem and/or to identify how the weakness of the methods can be reduced.

Probably one of the most important factors is the variability of the environmental conditions. Unlike the OpenAI-ES, the PPO algorithms include a mechanism for filtering out the noise affecting the reward measures caused by environmental variations. Such a mechanism consists in the calculation of the advantage which is based on an estimation of the expected reward. This enables the PPO to outperform the OpenAI-ES algorithm in problems in which the impact of environmental variations is large and in which the expected reward can be predicted with sufficient accuracy. It might enable the OpenAI-ES algorithm to outperform the PPO in problems in which predicting the expected reward is hard.

The weakness of OpenAI-ES in this respect can be reduced by using the super-symmetrical version of the algorithm introduced here, by extending the algorithm with a mechanism for estimating the relative difficulty of the environmental conditions and for selecting experience with a similar difficulty level (Milano and Nolfi, 2021), or , when possible, by altering the problem in a way which reduce the impact of environmental variation (see Section 6). The weakness of the PPO with respect to problems in which estimating the expected reward is difficult can be reduced by enriching the observation of the critic eventually with information which is available in simulation even if it cannot be accessed in hardware (see for example Andrychowicz et al., 2018).

A second important factor is the reward function. The OpenAI-ES algorithm operates effectively with reward functions which are implicit and sparse, i.e. which rewards the agent on the basis of its ability to achieve the desired goal only and in which the offset in time between the actions and the associated rewards is large. The PPO algorithm struggles more with reward functions which are implicit and sparse and benefit from the introduction of incentives. As demonstrated in Section 5, different reinforcement learning algorithms differ in this respect as well. These findings imply that the reward function should be optimized to the particular method used. Moreover, these findings imply that benchmarking comparison carried out without optimizing the reward functions provide little evidences on the relative efficacy of the compared methods.

A third important factor is the solution space which can be accessed by the two methods. The OpenAI-ES has access to a large set of solutions which include minimal solutions, i.e. solutions which operate on the basis of few control rules. The PPO instead has access to a restricted solution space which include the solutions capable to cope with large action perturbations only. This latter set of solutions might exclude minimal solutions. This qualitative difference represents a strength and a weakness for the OpenAI-ES and the PPO methods, respectively, for problems in which the simplicity of the solution is critical. Instead, it represents a weakness and a strength for the OpenAI-ES and the PPO methods, respectively, for problems in which minimal solutions correspond to local minima or in which the ability to cope with variations is crucial. These qualitative differences also explain why evolutionary and reinforcement learning methods often discover qualitatively different behavioral solutions.

Making the qualitative difference of alternative methods explicit also permit to clarify the relative weaknesses of each algorithm and to identify methods for ameliorating such weaknesses. For example, the realization of the inability of the OpenAI-ES method to deal with large environmental variations enable us to propose the super-symmetrical version of this algorithm which is more effective in this respect. Moreover, understanding the characteristics of the particular method used can be used to tune the characteristics of the experimental setup in a more informed way.

**Appendix**

Where possible, the hyperparameters were set by using the values included in Salimans et al. (2017) for the experiments performed with the OpenAI-ES algorithm and in Stable Baseline Zoo (https://github.com/araffin/rl-baselines-zoo) for the experiments performed with the PPO algorithm.

More specifically, in the case of the experiments performed with the OpenAI-ES: (i) the policy of the agents consisted of a 3-layers feedforward neural network with 64 internal neurons in all environments with the exception of the Atari problems for which we used a Convolutionary Architecture (Mnih et al. 2015). (ii) the agents were evaluated for 10 episodes, in the case of the BipedalWalkerHardcore environment, and for 1 episode, in the case of the other environments, (iii) the size of the population was set to 500 in the case of the Humanoid environment and to 40 in the case of the other environments, (iv) the evolutionary process was continued for a total of $2^6$ steps in the case of the Hopper and Walker2d and 1e7 steps in the case of the Atari, Humanoid, Bipedal Hardcore, and Slimevolley environments, (v) the observation vectors were normalized by using virtual batch normalization (Salimans et al, 2016, 2017), (vi) the connections weights were normalized by using weight decay, (vii) the distribution of the perturbations of parameters was set to 0.02, and (vii) the step size of the Adam optimizer was set to 0.01.

In the case of the experiments performed with the PPO: (i) the policy of the agents consisted of 4-layers feedforward neural networks with two layers of 256 internal neurons (in the case of the Hopper and Walker2D environments) and 256 internal neurons (in the case of the BipedalHardcore, Humanoid, and Slimevolley environments). For the Atari problems we used the standard convolutionary architecture (Mnih et al., 2015) (ii) the training process was continued for $2^6$ steps in the case of the Hopper and Walker2d and $1^7$ steps in the case of the Atari, Humanoid, Bipedal Hardcore, and Slimevolley environments, (iii) the rollout consists of 512, 512, 128, 2048 steps in the (1) Hopper, Walker2D, and (2) Slimevolley, (3) Atari, and (4) Humanoid and BipedalWalker environments, respectively, and was divided into minibatches of dimension 128, 32, 256, 64, respectively (iv) the observation vectors were normalized by using batch normalization, (v) the connections weights were normalized by using weight decay, (vi) the learning rate was increased linearly in the range [0.0, 0.004] in the case of the Hopper, Walker2d and Humanoid environments and was set to $2.5^{-4}$ in the case of the other environments, (vii) the threshold for clipping the gradient was set to 0.2 for the actor network and to 0.5 for the critic network, (viii) the entropy coefficient was set to 0.01 in the Atari environments and to 0.0 in the other environments.

**Video 1. https://youtu.be/hhY4pF9iO3U Behavior displayed by a typical agent trained with the OpenAI-ES algorithm on the Centipede Atari problem.**

**Video 2. https://youtu.be/s6B8UHZhfL8 Behavior displayed by a typical agent trained with the PPO algorithm on the Centipede Atari problem.**

**Video 3 https://youtu.be/3GQqGJIaOqg Behavior displayed by a typical agent trained with the OpenAI-ES algorithm on the Breakout Atari problem.**

**Video 4. https://youtu.be/GrLBXi57XIo Behavior displayed by a typical agent trained with the PPO algorithm on the Breakout Atari problem.**

**Video 5. https://youtu.be/S2NM-dkMhfA Behavior displayed by a typical agent trained with the OpenAI-ES algorithm on the PyBullet Hopper problem with the standard reward, including bonus to stay up.**

**Video 6. https://youtu.be/XfG6ltqZMAQ Behavior displayed by a typical agent trained with the PPO algorithm on the PyBullet Hopper problem with the standard reward, including bonus to stay up.**

**Video 7. https://youtu.be/uq6LuTOTgDQ Behavior displayed by a typical agent trained with the OpenAI-ES algorithm on the PyBullet Walker2d problem with the standard reward, including bonus to stay up.**

**Video 8. https://youtu.be/sUW0VaGZ1F0 Behavior displayed by a typical agent trained with the PPO algorithm on the PyBullet Walker2d problem with the standard reward, including bonus to stay up.**

**Video 9. https://youtu.be/PT63tHgdJ3o Behavior displayed by a typical agent trained with the OpenAI-ES algorithm on the PyBullet Hopper problem with the modified reward, without the bonus to stay up.**

**Video 10. https://youtu.be/FVnZvqoDipU Behavior displayed by a typical agent trained with the PPO algorithm on the PyBullet Hopper problem with the modified reward, without the bonus to stay up.**

**Video 11.** https://youtu.be/SsvOODQmxjE Behavior displayed by a typical agent trained with the OpenAI-ES algorithm on the PyBullet Walker2d problem with the modified reward, without the bonus to stay up.

**Video 12.** https://youtu.be/0p92vzBD5zg Behavior displayed by a typical agent trained with the PPO algorithm on the PyBullet Walker2d problem with the modified reward, without the bonus to stay up.